\documentclass{article}
\usepackage{spconf,amsmath,graphicx}
\usepackage{amssymb} % For mathematical symbols
\usepackage{hyperref} % For citations/links
\usepackage{booktabs} % For better tables
\usepackage{multirow} % For multirow tables
\usepackage{enumitem}
\setlist{nosep, leftmargin=14pt}

\usepackage{mwe} % to get dummy images (used for visualization placeholder)

\title{
LoSA-Net: A Localized and Scale-Adaptive Network for \\
Boundary-Sensitive Prediction of Perineural Invasion in 3D MRI
}

\name{%
\begin{tabular}{@{}c@{}}
Youngung Han$^{1,2}$, Hyunsu Go$^{1}$, Kyeonghun Kim$^{2}$,
Induk Um$^{3}$, Junga Kim$^{1}$, Jaewon Jung$^{1}$, \\
Woo Kyoung Jeong$^{4}$, Won Jae Lee$^{5}$,
Pa Hong$^{5}$, Ken Ying-Kai Liao$^{6}$,
Hyuk-Jae Lee$^{1}$, Nam-Joon Kim$^{1,\dagger}$
\end{tabular}
}

\address{
$^{1}$Seoul National University, Seoul, Republic of Korea \\
$^{2}$OUTTA, Seoul, Republic of Korea \\
$^{3}$Chung-Ang University, Seoul, Republic of Korea \\
$^{4}$Samsung Medical Center, Sungkyunkwan University School of Medicine, Seoul, Republic of Korea \\
$^{5}$Samsung Changwon Hospital, Changwon, Republic of Korea \\
$^{6}$NVIDIA AI Technology Center, Taipei, Taiwan \\[0.3em]
$^{\dagger}$Corresponding author: \texttt{knj01@snu.ac.kr}
}

\begin{document}
\ninept

\maketitle

\begin{abstract}
Perineural invasion (PNI) is a clinically relevant indicator of tumor aggressiveness and can influence surgical decision-making, motivating interest in reliable preoperative assessment. The subtle MRI features of PNI, however, often resemble nearby anatomy, complicating noninvasive prediction. These fine perineural cues are easily attenuated by routine downsampling or overly global feature aggregation, reducing the effectiveness of conventional volumetric models.
We present LoSA-Net, a localized and scale-adaptive architecture for boundary-sensitive PNI prediction in 3D MRI. Talking Neighborhood Attention (TNA) preserves nerve-aligned detail through localized self-attention with head-wise mixing, and Scale-Adaptive Feature Mixing (SAFM) modulates the receptive field using multi-scale depthwise processing. Cross-Scale Refinement and Alignment (CSRA) maintains consistency between semantic context and high-resolution boundaries across stages.
In contrast-enhanced MRI scans from 168 patients with cholangiocarcinoma, LoSA-Net achieves an AUC of 0.7567 and outperforms representative convolutional and transformer baselines under matched preprocessing and optimization settings. 
\end{abstract}

\section{Introduction}
\label{sec:intro}

% ------------------ Clinical background
Perineural invasion (PNI) is the spread of tumor cells along or within the nerve sheath and is associated with pain, neurologic deficits, and poor survival across multiple malignancies~\cite{Liebig2009PNI, Bakst2019PNI}. 
% It is recognized as an adverse prognostic factor in head and neck cancer, cutaneous squamous cell carcinoma, and cholangiocarcinoma~\cite{Tao2024HNSCC, Tu2023RG, Abdullaeva2024Diag}. 
PNI may warrant more aggressive margin considerations during surgery~\cite{Liu2024IJS, tang2016influence}, suggesting that reliable noninvasive preoperative identification could help avoid inappropriate treatment and support oncologically safer, tailored care. Despite its clinical relevance, PNI typically appears on MRI as faint, elongated signal changes or subtle perineural thickening near nerve-caliber structures, often resembling adjacent vessels or ducts and sometimes falling below the effective resolution of routine protocols~\cite{Tu2023RG, Abdullaeva2024Diag}. Consequently, preoperative detection remains challenging even for expert readers, and confirmation frequently relies on postoperative pathology.

% ---------------- Current computational approaches and limitations
Radiomics-based approaches have demonstrated that MRI carries useful information for noninvasive PNI prediction~\cite{Liu2024IJS, Zhang2023ProstatePNI, Sun2022PCCA}, but their hand-crafted features are sensitive to acquisition and scanner variability, which hampers reproducibility and external generalization~\cite{Mali2021Repro, Park2019Repro, Lee2024Harmonization}. These challenges motivate end-to-end volumetric models that learn features directly from the data.

Generic 3D convolutional and transformer backbones, however, have inductive biases that are not particularly aligned with the characteristics of PNI. 
Convolutional encoders~\cite{He2016ResNet, Huang2017DenseNet, Tan2019EfficientNet} achieve invariance by striding and pooling, which can diminish the thin perineural edges that differentiate PNI from look-alike anatomy.
Studies of U-Net variants and recent segmentation models highlight consistent difficulties at low-contrast, fine boundaries, reinforcing the need for boundary-aware, multi-scale designs~\cite{UNetReview2025, Srep2024DRANet}. 
Vision transformers provide long-range context but lack built-in locality~\cite{Khan2022TransSurvey, Shamshad2023MedTrans, Li2023TransformersMed}. 
Hierarchical or windowed designs~\cite{Liu2021Swin, Hassani2023NAT} recover local structure, although cross-scale fusion at stage transitions can still degrade weak boundary signals if not carefully constrained.

Recent hybrid architectures combine convolution and attention to better balance local detail and global context.
Multi-branch designs improve the balance between fine detail and broader context~\cite{Szegedy2015Inception, Xie2017ResNeXt}. 
Selective kernel mechanisms~\cite{Hu2018SENet, Li2019SKNet} adapt receptive fields to object scale, and depthwise separable convolutions with inverted bottlenecks~\cite{Chollet2017Xception, Sandler2018MobileNetV2} provide efficient multi-scale feature extraction. 
These trends suggest that locality, scale adaptation, and efficient depthwise processing are beneficial when modeling small, low-contrast structures such as perineural tracts.

Motivated by these observations, we frame PNI prediction in 3D MRI as a boundary-centric task that requires accurate modeling of nerve-aligned edges alongside stable multi-scale context. 
We propose LoSA-Net, a localized and scale-adaptive volumetric encoder that integrates Talking Neighborhood Attention (TNA), Scale-Adaptive Feature Mixing (SAFM), and Cross-Scale Refinement and Alignment (CSRA) to capture weak perineural signals better.
% We evaluate LoSA-Net on a cohort of patients with PNI status and compare its performance to representative convolutional and transformer baselines under matched training and preprocessing conditions.
\section{Methodology}
\label{sec:method}

\begin{figure*}[t]
\centering
\includegraphics[width=0.8\textwidth]{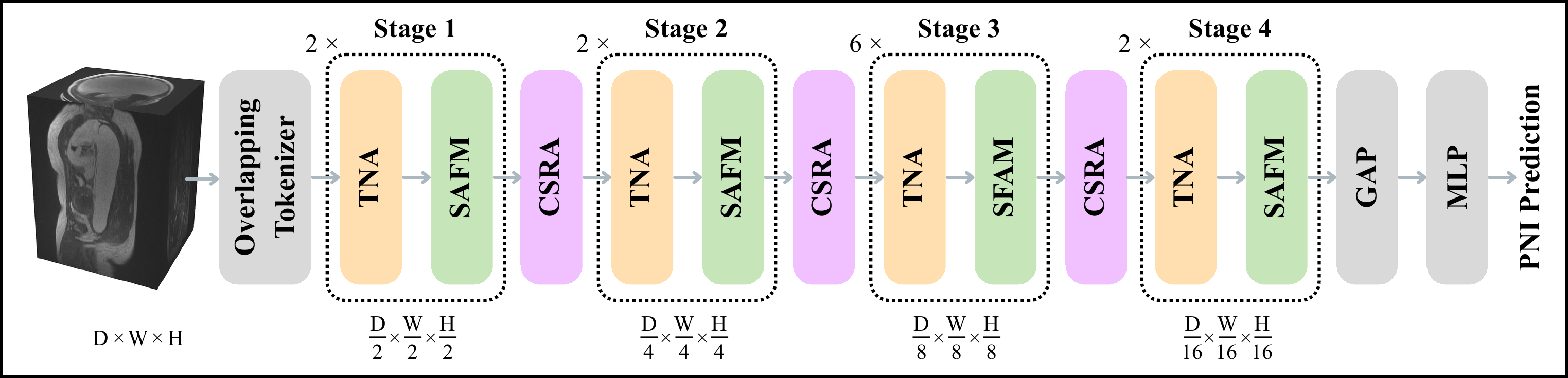}
\caption{
Overview of LoSA-Net. The network receives a tumor-centered crop and passes it through an overlapping convolutional tokenizer followed by four hierarchical stages. Each stage stacks blocks that combine Talking Neighborhood Attention (TNA) and Scale-Adaptive Feature Mixing (SAFM), while Cross-Scale Refinement and Alignment (CSRA) couples adjacent stages. Global average pooling and a linear classifier yield the final PNI prediction.
}
\label{fig:architecture_overview}
\end{figure*}

LoSA-Net is a four-stage volumetric encoder (Fig.~\ref{fig:architecture_overview}).
An overlapping convolutional tokenizer maps the input into a dense grid of tokens.
Each stage stacks blocks that couple Talking Neighborhood Attention (TNA) with Scale-Adaptive Feature Mixing (SAFM).
Cross-Scale Refinement and Alignment (CSRA) connects adjacent stages to maintain consistency between coarse semantic context and fine boundary features.
Global average pooling and a linear classifier produce the PNI prediction.

% ==========================================================

\subsection{Talking Neighborhood Attention (TNA)}
\label{ssec:tna}

TNA is designed to enhance nerve-aligned boundaries that are often indistinct in volumetric imaging by combining localized self-attention with head-wise mixing.

\begin{figure}[t]
\centering
\includegraphics[width=0.95\linewidth]{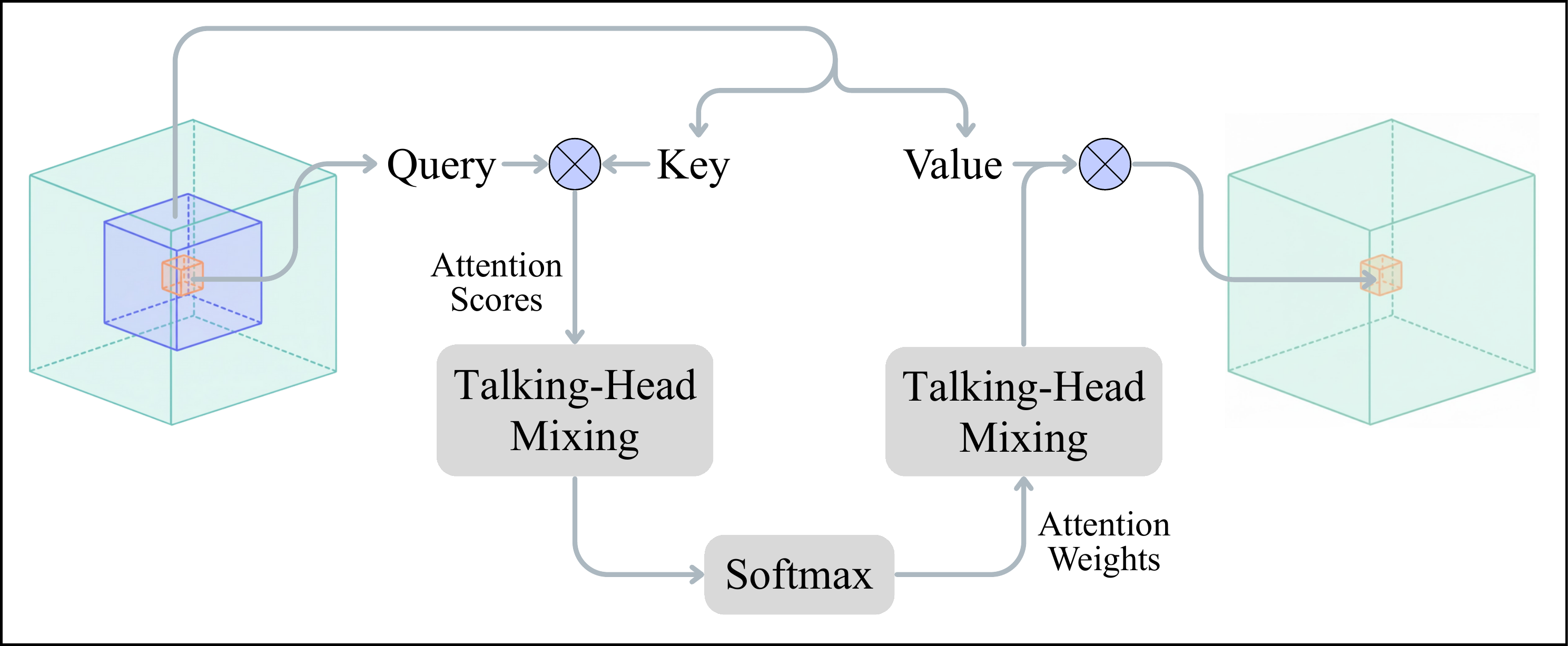}
\caption{Talking Neighborhood Attention (TNA).
Attention is restricted to a local 3D neighborhood to preserve spatial continuity.
Talking-head mixing aggregates directional cues across heads and stabilizes the response in heterogeneous tissue.}
\label{fig:tna_block}
\end{figure}

\paragraph*{Localized self-attention}
Given a volumetric feature map $\mathbf{X}$, each head projects it into queries, keys, and values:
\begin{equation}
\mathbf{Q}_h = \mathbf{W}^Q_h \mathbf{X},\quad
\mathbf{K}_h = \mathbf{W}^K_h \mathbf{X},\quad
\mathbf{V}_h = \mathbf{W}^V_h \mathbf{X},
\end{equation}
where $\mathbf{W}^Q_h,\mathbf{W}^K_h,\mathbf{W}^V_h$ are learnable projections shared across spatial locations and $h$ indexes the heads.
For each token index $i$, attention is restricted to a spatial neighborhood $\mathcal{N}(i)$ in the spirit of neighborhood attention~\cite{Hassani2023NAT}:
\begin{equation}
L_h(i,j) = \frac{\mathbf{Q}_{h,i}^{\top}\mathbf{K}_{h,j}}{\sqrt{d_h}},
\quad j\in\mathcal{N}(i).
\label{eq:tna_logit}
\end{equation}
This local region helps preserve continuity along perineural trajectories and limits spurious long-range mixing in unrelated parenchyma.

\paragraph*{Talking-head mixing}
Perineural structures exhibit directional variability. To capture these orientation-dependent cues, we apply talking-head mixing before and after the softmax~\cite{Shazeer2020TalkingHeads}.
% Perineural trajectories follow anisotropic orientations in three dimensions.
% To aggregate directional evidence across heads, we adopt talking-head style mixing before and after the softmax~\cite{Shazeer2020TalkingHeads}.
Let $\mathbf{L}(i,j)$ collect the scores $L_h(i,j)$ across attention heads.
A learnable pre-softmax mixing matrix $\mathbf{M}_\text{pre}$ linearly combines head-specific scores:
\begin{equation}
\widehat{\mathbf{L}}(i,j) = \mathbf{M}_{\text{pre}}\,\mathbf{L}(i,j).
\end{equation}
Head-wise attention weights are then obtained as:
\begin{equation}
A_h(i,j) = \mathrm{softmax}\big(\widehat{L}_h(i,j)\big).
\end{equation}
Another learnable matrix $\mathbf{M}_{\text{post}}$ mixes the normalized weights:
\begin{equation}
\widehat{\mathbf{A}}(i,j) = \mathbf{M}_{\text{post}}\,\mathbf{A}(i,j),
\end{equation}
and value aggregation follows:
\begin{equation}
\widetilde{\mathbf{O}}_h(i) = \sum_{j\in\mathcal{N}(i)} \widehat{A}_h(i,j)\,\mathbf{V}_{h,j}.
\end{equation}
This formulation gives LoSA-Net a controllable trade-off between local continuity and multi-directional context.

% ==========================================================
\subsection{Scale-Adaptive Feature Mixing (SAFM)}
SAFM adjusts the effective receptive field to capture both boundary detail and contextual patterns, using an inverted bottleneck with multi-scale depthwise convolutions and a lightweight scale selector.

\begin{figure}[t]
\centering
\includegraphics[width=0.95\linewidth]{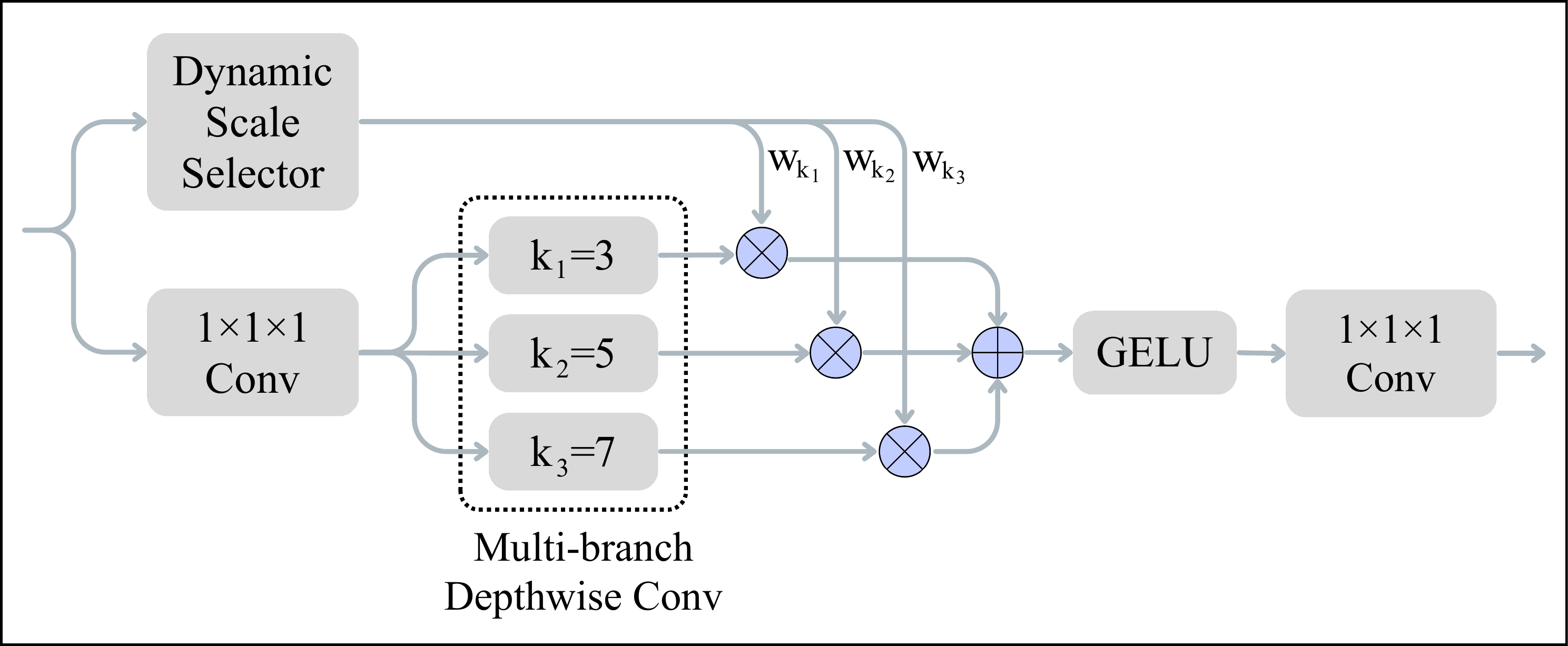}
\caption{Scale-Adaptive Feature Mixing (SAFM).
Parallel depthwise convolution branches with different kernel sizes provide complementary receptive fields.
A dynamic selector assigns fusion weights, allowing for the adaptive adjustment of the effective receptive field.}
\label{fig:safm_block}
\end{figure}

\paragraph*{Multi-scale depthwise processing}
SAFM begins by expanding channel capacity with a pointwise convolution:
\begin{equation}
\mathbf{U}=\mathbf{W}^{\uparrow} * \mathbf{X},
\end{equation}
where $*$ denotes convolution. 
Depthwise convolutions parameterized by spatial kernels $\mathbf{W}_k$ of size $k \times k \times k \in \mathcal{K}$ are applied:
\begin{equation}
\mathbf{Z}_k = \mathbf{W}_k *_{\mathrm{dw}} \mathbf{U}.
\end{equation}
Smaller kernels emphasize high-frequency boundary cues, while larger kernels capture broader context surrounding the tumor and perineural tracts.

\paragraph*{Dynamic scale selection}
A global descriptor:
\begin{equation}
\mathbf{c} = \mathrm{GAP}(\mathbf{X})
\end{equation}
conditions a lightweight MLP that predicts a set of fusion weights:
\begin{equation}
\mathbf{s} = \mathrm{softmax}\big(\mathrm{MLP}(\mathbf{c})\big).
\end{equation}
The aggregated output is projected back to the original channel dimension with a pointwise convolution:
\begin{equation}
\mathbf{Y} = \mathbf{W}^{\downarrow} * \phi\!\left(\sum_{k}s_k\,\mathbf{Z}_k\right),
\end{equation}
where $\phi$ is GELU.
% This mechanism allows the model to balance boundary-sensitive and context-sensitive information adaptively.
This mechanism allows the model to balance fine detail and contextual information adaptively.

% ==========================================================

\subsection{Cross-Scale Refinement and Alignment (CSRA)}
CSRA aligns multi-stage representations via complementary coarse-to-fine and fine-to-coarse pathways, with zero-initialized learnable scalars ensuring stable cross-scale interactions.
% CSRA aligns representations across stages through complementary coarse-to-fine and fine-to-coarse pathways. Zero-initialized learnable scalars modulate these paths to ensure stable cross-scale interactions.

\begin{figure}[t]
\centering
\includegraphics[width=0.95\linewidth]{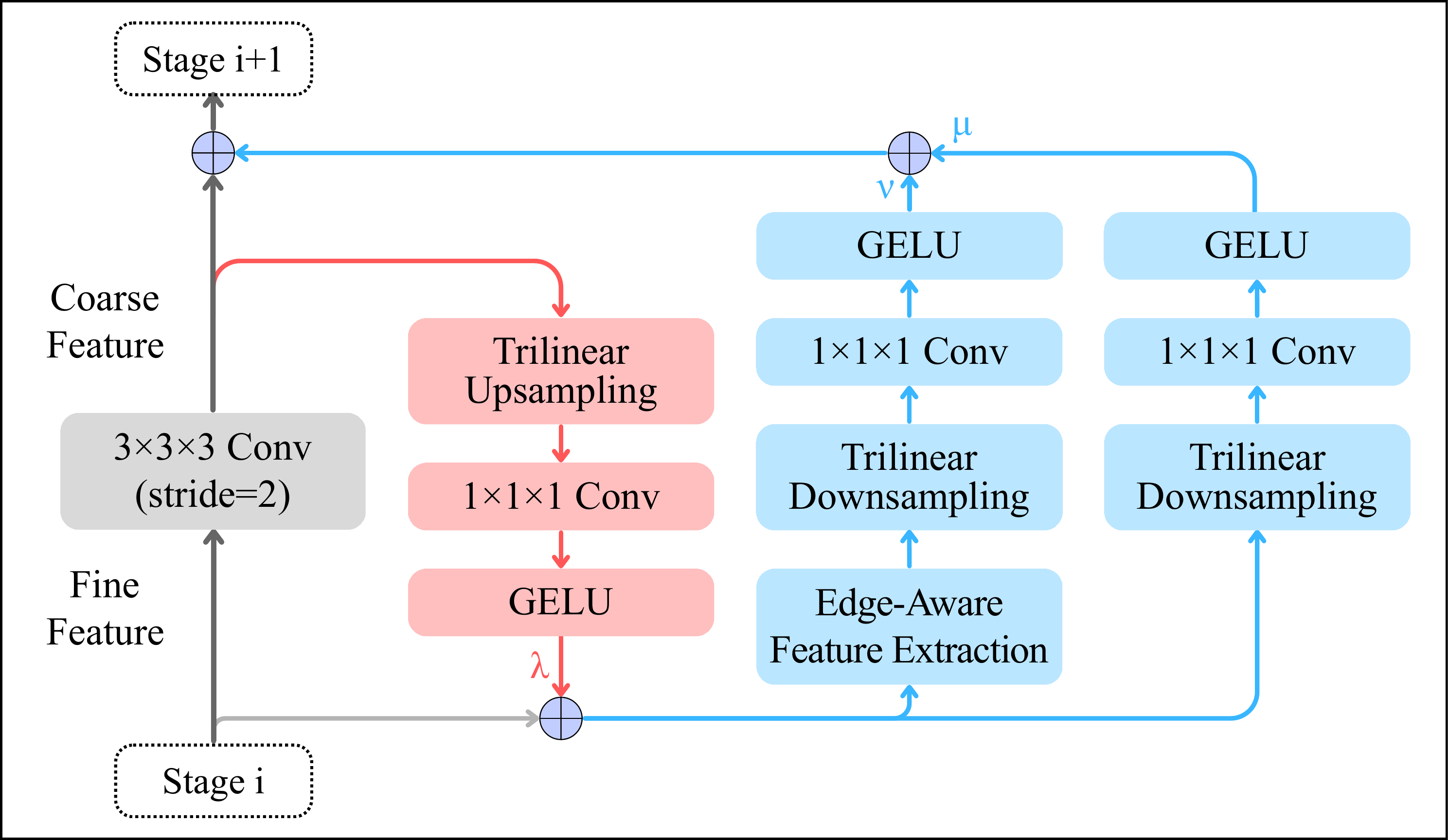}
\caption{Cross-Scale Refinement and Alignment (CSRA).
A coarse-to-fine path (red) injects semantic context, while a fine-to-coarse path (blue) propagates boundary cues using a Laplacian-based edge prior with a downsampled feature stream.
Zero-initialized scalars ($\lambda$, $\mu$, $\nu$) regulate cross-scale interactions for stable training.
% A coarse-to-fine path (red) injects semantic context, while a fine-to-coarse path (blue) comprises both a downsampled feature stream and a Laplacian-based edge prior, enabling
% the propagation of boundary information to deeper stages.
% Zero-initialized learnable scalars ($\lambda$, $\mu$, $\nu$) govern cross-scale interactions for stable training.
}
\label{fig:csra_block}
\end{figure}

\paragraph*{Stage transition}
Downsampling from stage $\ell$ to $\ell+1$ is performed using a strided convolution:
\begin{equation}
\mathbf{F}_{\ell+1}' = \mathbf{W}_\ell *_{\text{stride}} \mathbf{F}_\ell,
\end{equation}
with partial overlap to reduce aliasing of thin boundary patterns.

\paragraph*{Coarse-to-fine semantic injection}
Coarse features are upsampled through trilinear interpolation $\mathcal{U}$ and injected into the finer stage via a gated residual path:
\begin{equation}
\widehat{\mathbf{F}}_\ell
= \mathbf{F}_\ell
+ \lambda_\ell\,\phi\!\left(\mathbf{W}^{\lambda}_\ell * \mathcal{U}(\mathbf{F}_{\ell+1}^{\prime})\right),
\end{equation}
where $\lambda_\ell$ is a learnable scalar initialized to zero.

\paragraph*{Fine-to-coarse boundary feedback}
Fine-scale features contain boundary cues that are absent after downsampling.  
CSRA extracts a lightweight edge prior using a discrete 3D Laplacian:
\begin{equation}
\mathbf{E}_\ell = \mathcal{L}\bigl(\mathrm{Mean}_c(\widehat{\mathbf{F}}_\ell)\bigr),
\end{equation}
where $\mathrm{Mean}_c$ denotes channel averaging and $\mathcal{L}$ is a discrete 3D Laplacian operator.
Both $\widehat{\mathbf{F}}_\ell$ and $\mathbf{E}_\ell$ are downsampled through trilinear interpolation $\mathcal{D}$ and injected into the coarse stream:
\begin{equation}
\widetilde{\mathbf{F}}_{\ell+1}
= \mathbf{F}_{\ell+1}'
+ \mu_\ell\,\phi\!\left(\mathbf{W}^{\mu}_\ell * \mathcal{D}(\widehat{\mathbf{F}}_\ell)\right)
+ \nu_\ell\,\phi\!\left(\mathbf{W}^{\nu}_\ell * \mathcal{D}(\mathbf{E}_\ell)\right),
\end{equation}
with $\mu_\ell$ and $\nu_\ell$ initialized to zero.
This mechanism aligns coarse semantic representations with fine boundary information while maintaining stable optimization.

\section{Experiments and Results}
\label{sec:experiments}

% ------------------ 11/10
\subsection{Dataset and Implementation}
We utilized contrast-enhanced, hepatobiliary-phase T1-weighted MRI volumes from Samsung Medical Center, acquired over approximately 10 years. From 306 initial candidates, scans with severe motion artifacts, missing sequences, incorrect labels, or tumor burden exceeding 70\% of liver volume were excluded. The final cohort consisted of 168 patients with pathologically confirmed cholangiocarcinoma, including 67 PNI-positive and 101 PNI-negative cases. 

All volumes were converted to NIfTI, resampled to a unified voxel grid of $96 \times 96 \times 48$, and z-score normalized. Tumor-centered crops were extracted to emphasize peri-tumoral regions where imaging correlates of perineural spread are expected. Data splitting was performed at the patient level using 5-fold cross-validation, and all results are reported as the mean performance across folds.

LoSA-Net was implemented in PyTorch with mixed precision. TNA operated on 3D neighborhoods using the efficient NATTEN kernel~\cite{Hassani2023NAT}. The SAFM module employed parallel depthwise branches with kernel sizes $\mathcal{K}=\{3,5,7\}$. We used AdamW~\cite{Loshchilov2017AdamW} with an initial learning rate of $1{\times}10^{-4}$ and cosine decay. Models were trained for 200 epochs with a batch size of 8.

Class imbalance was addressed using class-balanced focal loss~\cite{Lin2017FocalLoss} with dataset-derived weighting factor $\alpha$ and a focusing parameter $\gamma=2$. Spatial augmentations were restricted to mild rotations, flips, and intensity jitter to maintain anatomical plausibility.

% ----------------- 11/10
\subsection{Comparison with State-of-the-Art Models}
\begin{table}[t]
\centering
\caption{Performance comparison of representative 3D models on the PNI cohort.}
\label{tab:sota}
\begin{tabular}{@{}llc@{}}
\toprule
\textbf{Category} & \textbf{Model (3D)} & \textbf{AUC} \\ 
\midrule
CNN 
& ResNet~\cite{He2016ResNet}         & 0.6951 \\
& DenseNet~\cite{Huang2017DenseNet}   & 0.6787 \\
& EfficientNet~\cite{Tan2019EfficientNet} & 0.6696 \\
\midrule
Transformer
& Swin Transformer~\cite{Liu2021Swin} & 0.6822 \\
& Neighborhood Attention Transformer~\cite{Hassani2023NAT}           & 0.6835 \\
\midrule
\textbf{Hybrid}
& \textbf{LoSA-Net (Ours)} & \textbf{0.7567} \\
\bottomrule
\end{tabular}
\end{table}

% \begin{table}[t]
% \caption{Performance of representative backbones on the PNI cohort.}
% \label{tab:sota}
% \begin{tabular*}{\linewidth}{@{\extracolsep{\fill}}lll}
% \toprule
% Model & Type & AUC \\
% \midrule
% ResNet & CNN & 0.6951 \\
% DenseNet & CNN & 0.6787 \\
% EfficientNet & CNN & 0.6696 \\
% Swin Transformer & Transformer & 0.6822 \\
% NAT & Transformer & 0.6835 \\
% \midrule
% \textbf{LoSA-Net (Ours)} & Hybrid & \textbf{0.7567} \\
% \bottomrule
% \end{tabular*}
% \end{table}

We compared LoSA-Net against representative 3D convolutional and transformer-based models under identical preprocessing and optimization settings.
As summarized in Table~\ref{tab:sota}, LoSA-Net achieved the highest AUC of 0.7567 on the PNI cohort, outperforming both convolutional and transformer baselines under matched training setups. Convolutional backbones exhibited lower discrimination, which we attribute to their stronger emphasis on coarse patterns after repeated striding and pooling. Transformer variants benefited from long-range context but showed limited gains, consistent with the challenges of preserving thin boundaries across stage transitions.

% ------------------- 11/10
\subsection{Ablation Study}
\label{ssec:ablation}

We evaluated the contribution of each component of LoSA-Net under identical training and data splits. 
For the CSRA initialization experiment, the cross-scale scalars $(\lambda_\ell, \mu_\ell, \nu_\ell)$ were initialized either to zero (default) or sampled from $\mathcal{N}(0, \sigma^{2})$ with $\sigma \in \{0.1, 0.5\}$.

\begin{table}[t]
\caption{Ablation results on the PNI cohort. Each setting isolates the effect of a single architectural choice.}
\label{tab:ablation}
\begin{tabular*}{\linewidth}{@{\extracolsep{\fill}}lll}
\toprule
\textbf{Category} & \textbf{Configuration} & \textbf{AUC} \\
\midrule
\textbf{Input} & Uncropped Input & 0.6493 \\
\midrule
\multirow{3}{*}{\textbf{Module}} 
 & w/o TNA & 0.7085 \\
 & w/o SAFM & 0.7004 \\
 & w/o CSRA & 0.7209 \\
\midrule
\multirow{4}{*}{\textbf{SAFM Variants}} 
 & Multi-Branch w/o Scale Selector & 0.7364 \\
 & Single-Branch $\mathcal{K}{=}\{3\}$
 & 0.7143 \\
 & Single-Branch $\mathcal{K}{=}\{5\}$ & 0.7169 \\
 & Single-Branch $\mathcal{K}{=}\{7\}$
 & 0.7225 \\
\midrule
\multirow{2}{*}{\textbf{CSRA Initialization}} 
 & $\sigma=0.1$ & 0.7497 \\
 & $\sigma=0.5$ & 0.7301 \\
\midrule
\textbf{Baseline} & \textbf{LoSA-Net (Ours)} & \textbf{0.7567} \\
\bottomrule
\end{tabular*}
\end{table}

The results in Table~\ref{tab:ablation} show that each module contributes to overall performance. 
Removing TNA or SAFM degrades discrimination, indicating that localized attention and multi-scale depthwise processing provide complementary information.
Eliminating CSRA reduces AUC, highlighting the importance of cross-scale alignment for maintaining coherent representations.
For SAFM, multi-branch processing remains beneficial even without the selector, while adaptive fusion yields the best results.
For CSRA, zero initialization leads to more stable optimization than sampling with $\sigma > 0$.

\subsection{Qualitative Analysis}

\begin{figure}[t]
\centering
% Placeholder for Grad-CAM visualization (using mwe example-image)
\includegraphics[width=0.97\linewidth]{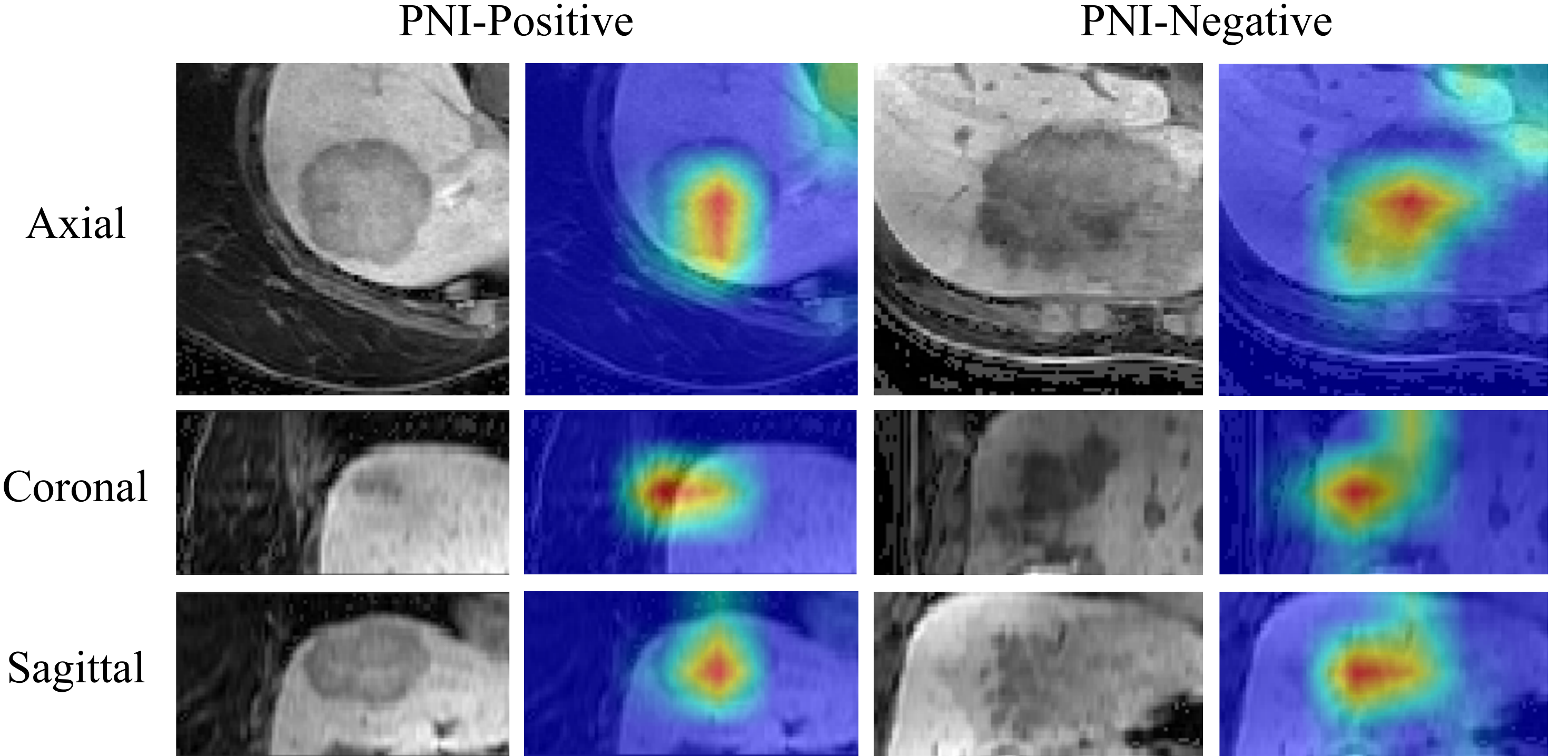}
\caption{Grad-CAM overlays for representative PNI-positive and PNI-negative cases in axial, coronal, and sagittal views. Heat maps are normalized per volume and overlaid on T1-weighted MRI.}
\label{fig:gradcam}
\end{figure}

3D Grad-CAM~\cite{Selvaraju2017GradCAM} maps were generated from the final stage of LoSA-Net (Fig.~\ref{fig:gradcam}). 
The visualizations showed increased responses near tumor boundaries and along adjacent tubular structures, which is consistent with the model’s boundary-oriented design.

% 3D Grad-CAM maps~\cite{Selvaraju2017GradCAM} were generated from the final stage of LoSA-Net (Fig. \ref{fig:gradcam}). 
% The patterns confirm that the model focuses on the tumor boundaries and adjacent nerve pathways, aligning with the intended boundary-sensitive design.
% \input{4_disc}
\section{Discussion and Conclusion}
\label{sec:conclusion}
This study addressed the problem of predicting PNI in 3D MRI, where the relevant imaging patterns are subtle, elongated, and often obscured by nearby vascular or ductal structures. Across a cohort of patients with cholangiocarcinoma, LoSA-Net achieved higher discrimination than representative convolutional and transformer models, indicating that boundary-sensitive modeling can provide advantages over generic volumetric architectures.

LoSA-Net incorporates inductive biases that reflect the imaging characteristics of PNI. TNA confines self-attention to localized 3D neighborhoods and mixes directional cues across heads, aiming to capture continuity along nerve-caliber structures better. SAFM adjusts the effective receptive field using multi-branch depthwise convolutions, enabling the encoder to emphasize boundary detail and surrounding context. CSRA promotes consistency between coarse semantic features and fine boundary cues across stages. Ablation experiments support the contribution of each module, and Grad-CAM visualizations show that the network focuses on boundary regions in a manner consistent with radiologic expectations.

This work has several limitations. The analysis is based on a single-center dataset with relatively homogeneous imaging protocols, and the sample size is modest. Tumor-centered crops focus on peri-tumoral regions but do not represent the full course of nerve pathways. PNI labels may include microscopic invasion that lies below the spatial resolution, constraining the degree to which MRI can fully describe the underlying ground truth.

% Future work should assess generalizability on multi-center datasets, explore multi sequence or multimodal extensions, and investigate methods that explicitly model nerve trajectories. Reader studies and calibration analyses may further clarify how boundary-aware models can be integrated into clinical workflows.

In summary, LoSA-Net offers a localized and scale-adaptive approach for boundary-sensitive prediction of PNI in 3D MRI. Its improvements over representative baseline models suggest that incorporating boundary- and scale-oriented inductive biases is a useful strategy for tasks involving small, low-contrast structures in volumetric medical imaging.

% We presented LoSA-Net, a localized and scale-adaptive volumetric encoder for boundary-sensitive prediction of perineural invasion in 3D MRI.
% The architecture incorporates localized attention, multi-scale depthwise feature mixing, and cross-scale refinement to enhance faint perineural boundaries while preserving multi-scale context.
% Across a cohort of cholangiocarcinoma patients, LoSA-Net outperformed representative convolutional and transformer backbones under identical training conditions.
% These findings suggest that boundary- and scale-aware inductive biases are valuable for tasks involving small, low-contrast structures in volumetric imaging.
% Future work will extend evaluation to multi-center datasets and explore integration with broader clinical workflows.
\section{Acknowledgments}
\label{sec:acknowledgments}

% This work was supported by the Institute of Information \& Communications Technology Planning \& Evaluation (IITP), funded by the Korea government (MSIT), under the Artificial Intelligence Semiconduct.

This work was supported by the Institute of Information \& Communications Technology Planning \& Evaluation (IITP), funded by the Korea government (MSIT), under the Artificial Intelligence Semiconductor Support Program to nurture the best talents (IITP-2023-RS-2023-00256081) and the grant for the Development of an AI Deep Learning Processor and Module for a 2,000 TFLOPS Server (No. 2020-0-01305).

% References should be produced using the bibtex program from suitable
% BiBTeX files. The IEEEbib.bst bibliography style file from IEEE produces
% unsorted bibliography list.
% -------------------------------------------------------------------------
\bibliographystyle{IEEEtran}
\bibliography{refs}

\end{document}